\documentclass{article}

\usepackage{times}
\usepackage{graphicx} 
\usepackage{subfigure} 

\usepackage{natbib}

\usepackage{amsmath, amsthm, amssymb}

\usepackage{enumitem}

\usepackage{lipsum}
\usepackage{ url }

\usepackage[ruled,lined,algo2e]{algorithm2e}

\newtheorem{definition}{Definition}

\newtheorem{thm}{Theorem}
\newtheorem{condition}{Condition}

\newenvironment{proofsketch}[1][\proofname]{\proof[#1]\mbox{}\\*}{\endproof}

\usepackage[nohyperref,accepted]{icml2015}

\usepackage{ dsfont }

\DeclareMathOperator*{\argmax}{arg\max}

\newcommand{\W}{\mathbf{W}}
\newcommand{\w}{\mathbf{w}}
\newcommand{\h}{\mathbf{h}}
\newcommand{\V}{\mathbf{V}}
\newcommand{\Hm}{\mathbf{H}}

\newcommand{\thetab}{{\boldsymbol \theta}}
\newcommand{\xib}{{\boldsymbol \zeta}}

\newcommand{\noisebw}{{\boldsymbol \Psi}}
\newcommand{\noisebh}{{\boldsymbol \Xi}}

\icmltitlerunning{Parallel Stochastic Gradient MCMC for Matrix Factorisation Models}

\begin{document} 

\twocolumn[
\icmltitle{Parallel Stochastic Gradient Markov Chain Monte Carlo for \\ Matrix Factorisation Models}

\icmlauthor{Umut \c Sim\c sekli$^\text{1}$}{umut.simsekli@boun.edu.tr}
\icmlauthor{Hazal Koptagel$^\text{1}$}{hazal.koptagel@boun.edu.tr}
\icmlauthor{Hakan G\"{u}lda\c s$^\text{1}$}{hakan.guldas@boun.edu.tr}
\icmlauthor{A. Taylan Cemgil$^\text{1}$}{taylan.cemgil@boun.edu.tr}
\icmlauthor{Figen \"{O}ztoprak$^\text{2}$}{figen.oztoprak@bilgi.edu.tr}
\icmlauthor{\c S. \.{I}lker Birbil$^\text{3}$}{sibirbil@sabanciuniv.edu}
\icmladdress{1: Department of Computer Engineering, Bo\u{g}azi\c ci University,  \.{I}stanbul, Turkey} \vspace{-10pt}
\icmladdress{2: Department of Industrial Engineering, Bilgi University, \.{I}stanbul, Turkey} \vspace{-10pt}
\icmladdress{3: Faculty of Engineering and Natural Sciences, Sabanc\i{} University,  \.{I}stanbul, Turkey}

\icmlkeywords{Stochastic gradient Langevin dynamics, Non-negative matrix factorisation, Distributed inference, Tweedie distribution}

\vskip 0.3in
]

\begin{abstract}  
For large matrix factorisation problems, we develop a distributed Markov Chain Monte Carlo (MCMC) method based on stochastic gradient Langevin dynamics (SGLD) that we call Parallel SGLD (PSGLD). PSGLD has very favourable scaling properties with increasing data size and is comparable in terms of computational requirements to optimisation methods based on stochastic gradient descent. PSGLD achieves high performance by exploiting the conditional independence structure of the MF models to sub-sample data in a systematic manner as to allow parallelisation and distributed computation. We provide a convergence proof of the algorithm and verify its superior performance on various architectures such as Graphics Processing Units, shared memory multi-core systems and multi-computer clusters. 
\end{abstract}

\section{Introduction}

Matrix factorisation (MF) models have been widely used in data analysis and have been shown to be useful in various domains, such as recommender systems, audio processing, finance, computer vision, and bioinformatics \cite{Smaragdis03,DevarajanBio,cichocki09}. The aim of a MF model is to decompose an observed data matrix $\V \in \mathds{R}^{I \times J}$ in the form: $\V \approx \W \Hm$, where $\W \in \mathds{R}^{I \times K}$ and $\Hm \in \mathds{R}^{K \times J}$ are the factor matrices, known typically as the dictionary and the weight matrix respectively, to be estimated by minimising some error measure such as the Frobenious norm $\|\V - \W \Hm\|_F$.

More general noise models and regularisation methods can be developed. One popular approach is using a probabilistic MF model
having the following hierarchical generative model:
\begin{gather}
\nonumber p(\W)  = \prod_{ik} p(w_{ik}) , \qquad
\nonumber p(\Hm) = \prod_{kj} p(h_{kj})
\end{gather}
\begin{gather}
p(\V|\W\Hm) = \prod_{ij} p(v_{ij}| \w_i,\h_j) \label{eqn:generative}
\end{gather}
where, $\w_i$ denotes the $i^{\text{th}}$ row of $\W$ and $\h_j$ denotes the $j^{\text{th}}$ column of $\Hm$\footnote{In the rest of the paper, we will use bold capital letters to denote matrices, e.g., $\mathbf{A}$, bold small letters to denote vectors, e.g., $\mathbf{a}$, and small regular letters to denote scalars, e.g., $a$.}.
In MF problems we might be interested in two different quantities:
\begin{enumerate}[noitemsep,topsep=0pt,leftmargin=*]
\item Point estimates such as the maximum likelihood (ML) or maximum a-posteriori (MAP):
\begin{align}
  (\W,\Hm)^\star &= \argmax_{\W,\Hm} \log p(\W,\Hm|\V) \label{eqn:map}
\end{align}
\item The full posterior distribution: 
\begin{align}
	p(\W,\Hm|\V) \propto p(\V|\W,\Hm) p(\W) p(\Hm) \label{eqn:fullbayes}
\end{align}
\end{enumerate}

The majority of the current literature on MF focuses on obtaining point estimates via optimisation of the objective given in Equation~\ref{eqn:map}. Point estimates can be useful in practical applications and there is a broad literature for solving this optimisation problem for a variety of choices of prior and likelihood functions, with various theoretical guarantees \cite{lee99,Liu2010,fevotte2011algorithms,gemulla2011,recht2013parallel}. In contrast, Monte Carlo methods that sample from the often intractable full posterior distribution (in the sense of computing moments or the normalizing constant) received less attention, mainly due to the perceived computational obstacles and rather slow convergence of standard methods, such as the Gibbs sampler, for the target density in \ref{eqn:fullbayes}.

Having an efficient sampler that can generate from the full posterior in contrast to a point estimate would be useful in various applications such as model selection (i.e., estimating the `rank' $K$ of the model) or estimating the Bayesian predictive densities useful for active learning. Yet, despite the well known advantages, Monte Carlo methods are typically not the method choice in large scale MF problems as they are perceived to be computationally very demanding. Indeed, classical approaches based on batch Metropolis-Hastings would require passing over the whole data set at each iteration and the acceptance step makes the methods even more impractical for large data sets. Recently, alternative approaches have been proposed to scale-up MCMC inference to large-scale regime. An important attempt was made by Welling and Teh \citeyearpar{WelTeh2011a}, where the authors combined the ideas from a gradient-based MCMC method, so called the Langevin dynamics (LD) \cite{neal2010} and the popular optimisation method, stochastic gradient descent (SGD) \cite{kushner}, and developed a scalable MCMC framework called as the stochastic gradient Langevin dynamics (SGLD). Unlike conventional batch MCMC methods, SGLD requires to `see' only a small subset of the data per iteration similar to SGD. With this manner, SGLD can handle large datasets while at the same time being a valid MCMC method that forms a Markov Chain asymptotically sampling from the target density. Approximation analysis of SGLD has been studied in \cite{icml2014c2_satoa14} and \cite{TehThiVol2014a}. Several extensions of SGLD have been proposed. Ahn et al. \citeyearpar{AhnKorWel2012} made use of the Fisher information besides the noisy gradients, Patterson and Teh \citeyearpar{PatTeh2013a} applied SGLD on the probability simplex. Chen et al. \citeyearpar{ChenICML2014} and Ding et al. \citeyearpar{DingFBCSN14} considered second order Langevin dynamics and made use of the momentum terms, extending the vanilla SGLD.

In this study, we develop a parallel and distributed MCMC method for sampling from the full posterior of a broad range of MF models, including models not easily tackled using standard methods such as the Gibbs sampler. Our approach is carefully designed for MF models and builds upon the generic distributed SGLD (DSGLD) framework that was proposed in \cite{AhnShaWel2014} where separate Markov chains are run in parallel on different subsets of the data that are distributed among worker nodes. When applied to MF models, DSGLD results in computational inefficiencies since it cannot exploit the conditional independence structure of the MF models. Besides, DSGLD requires all the latent variables (i.e., $\W$ and $\Hm$) to be synchronised once in a couple of iterations which introduces a significant amount of communication cost. On the other hand, for large problems it may not even be possible to store the latent variables in a single machine; one needs to distribute the latent variables among the nodes as well.  

We propose a novel parallel and distributed variant of SGLD for MF models, that we call Parallel SGLD (PSGLD). PSGLD has very favourable scaling properties with increasing data size, remarkably upto the point that the resulting sampler is computationally not much more demanding than an optimisation method such as the distributed stochastic gradient descent (DSGD) \cite{gemulla2011}. Reminisicent to DSGD, PSGLD achieves high performance by exploiting the conditional independence structure of the MF models for sub-sampling the data in a systematic manner as to allow parallelisation. The main advantages of PSGLD can be summarised as follows:
\begin{itemize}[noitemsep,topsep=0pt,leftmargin=*]
\item Due to its inherently parallel structure, PSGLD is faster than SGLD by several orders of magnitude while being as accurate.
\item As we will illustrate in our experiments, PSGLD can easily be implemented in both shared-memory and distributed architectures. This makes the method suitable for very large data sets that might be distributed among many nodes.
\item Unlike DSGLD, which requires to communicate all the parameters $\W$ and $\Hm$ among the worker nodes, PSGLD communicates only small parts of $\Hm$. This drastically reduces the communication cost for large $\W$ and $\Hm$. 
\item The probability distribution of the samples generated by PSGLD converges to the Bayesian posterior. 
\end{itemize}

We evaluate PSGLD on both synthetic and real datasets. Our experiments show that, PSGLD can be beneficial in two different settings: 1) a shared-memory setting, where we implement PSGLD on a graphics processing unit (GPU) 2) a distributed setting, where we implement PSGLD on a cluster of computers by using a message passing protocol. Our results show that, in the shared-memory setting, while achieving the same quality, PSGLD is $700+$ times faster than a Gibbs sampler on a non-negative matrix factorisation problem; and in the distributed setting, PSGLD easily scales-up to matrices with hundreds of millions of entries. 

We would like to note that, a DSGLD-based, distributed MF framework has been independently proposed by Ahn et al. \citeyearpar{Ahn15}, where the authors focus on a particular MF model, called as the Bayesian probabilistic matrix factorisation (BPMF) \cite{salakhutdinov2008bayesian}. In this study, we focus on a generalised observation model family (Tweedie models), in which we can obtain several observation models that have been used in important MF models (such as BPMF, Poisson non-negative matrix factorisation (NMF) \cite{lee99}, Itakura-Saito NMF \cite{cedric-neco09}) as special cases.

\begin{figure*}[t]
\begin{center}
\centerline{\includegraphics[width=1.75\columnwidth]{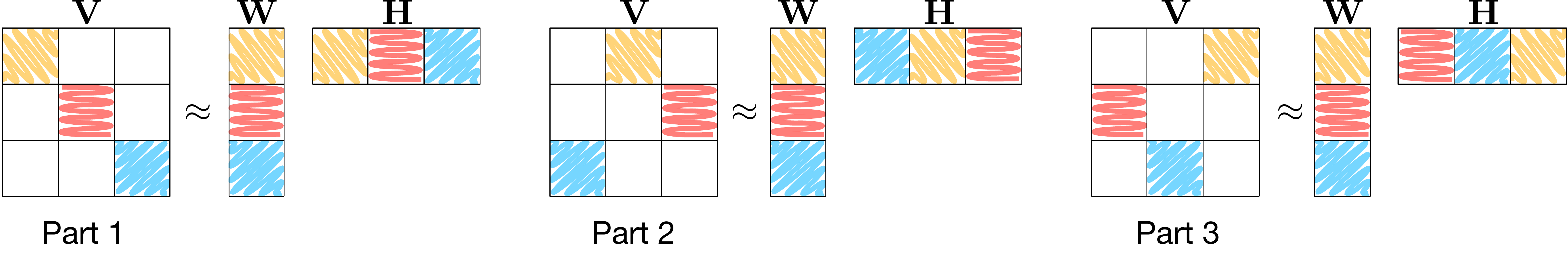}}
\vspace{-10pt}
\caption{Illustration of the \emph{parts} and the \emph{blocks}. Here, we partition the sets $[I]$ and $[J]$ into $B = 3$ pieces. The partitions for this example are chosen as ${\cal P}_B([I]) = \{ \{1,\dots, \frac{I}{3}\}, \{\frac{I}{3}+1,\dots, \frac{2I}{3}\}, \{\frac{2I}{3}+1,\dots, I \} \}$ and  ${\cal P}_B([J]) = \{ \{1,\dots, \frac{J}{3}\}, \{\frac{J}{3}+1,\dots, \frac{2J}{3}\}, \{\frac{2J}{3}+1,\dots, J \} \}$. Part 1 consists of three non-overlapping blocks, say $ \Pi = \Lambda_1 \cup \Lambda_2 \cup \Lambda_3$, where $\Lambda_1 = \{1,\dots, \frac{I}{3}\} \times \{1,\dots, \frac{J}{3}\}$, $\Lambda_2 = \{\frac{I}{3}+1,\dots, \frac{2I}{3}\}\times \{\frac{J}{3}+1,\dots, \frac{2J}{3}\}$, and $\Lambda_3 = \{\frac{2I}{3}+1,\dots, I\}\times \{\frac{2J}{3}+1,\dots,J\}$. Given the blocks in a part, the corresponding blocks in $\W$ and $\Hm$ become conditionally independent, as illustrated in different colours and textures. Therefore, for different blocks, the PSGLD updates can be applied in parallel.}
\label{fig:illus}
\end{center}
\vskip -0.2in
\end{figure*}

\section{Stochastic Gradient Langevin Dynamics (SGLD) for Matrix Factorisation}

In the last decade, SGD has become very popular due to its low computational requirements and convergence guarantee. SGLD brings the ideas of SGD and LD together in order to generate samples from the posterior distribution in a computationally efficient way. In algorithmic sense, SGLD is identical to SGD except that it injects a Gaussian noise at each iteration. For MF models, SGLD iteratively applies the following update rules in order to obtain the samples $\W^{(t)}$ and $\Hm^{(t)}$: $\W^{(t)} = \W^{(t-1)} + \Delta \W^{(t)}$ and $\Hm^{(t)} = \Hm^{(t-1)} + \Delta \Hm^{(t)}$, where
\begin{align*}
\nonumber \Delta \W^{(t)} =& \epsilon^{(t)} \Bigl( \frac{N}{|\Omega^{(t)}|} \hspace{-2pt} \sum_{(i,j)\in \Omega^{(t)}} \hspace{-10pt} \nabla_{\W}  \log p(v_{ij} | \W^{(t-1)},\Hm^{(t-1)})  \\ &
   + \nabla_{\W}  \log p(\W^{(t-1)})    \Bigr)  + \noisebw^{(t)}  \\
\nonumber \Delta \Hm^{(t)} =&  \epsilon^{(t)} \Bigl( \frac{N}{|\Omega^{(t)}|} \hspace{-2pt} \sum_{(i,j)\in \Omega^{(t)}} \hspace{-10pt} \nabla_{\Hm} \log p(v_{ij} | \W^{(t-1)},\Hm^{(t-1)}) \\ &
+  \nabla_{\Hm} \log p(\Hm^{(t-1)})    \Bigr)  + \noisebh^{(t)} .
\end{align*}
Here, $N$ is the number of elements in $\V$, $t = 1,\dots,T$ denotes the iteration number, $\Omega^{(t)} \subset [I] \times [J]$ is the sub-sample that is drawn at iteration $t$, the set $[I]$ is defined as $[I] = \{1,\dots, I\}$, $\nabla$ denotes the gradients, and $|\Omega^{(t)}|$ denotes the number of elements in $\Omega^{(t)}$. The elements of the noise matrices $\noisebw^{(t)}$ and $\noisebh^{(t)}$ are independently Gaussian distributed:
\begin{align*}
\psi_{ik}^{(t)} \sim {\cal N}(\psi_{ik}^{(t)}; 0, 2\epsilon^{(t)}), \quad \quad  \xi_{kj}^{(t)} \sim {\cal N}(\xi_{kj}^{(t)}; 0, 2\epsilon^{(t)}).
\end{align*}
For convergence, the step size $\epsilon^{(t)}$ must satisfy the following conditions:
\begin{align}
\sum_{t=0}^\infty \epsilon^{(t)} = \infty, \quad 
\sum_{t=0}^\infty (\epsilon^{(t)})^2 < \infty \label{eqn:eps}
\end{align}
A typical choice for the step size is $\epsilon^{(t)} = {\cal O}(1/t)$.

In SGLD, the sub-sample $\Omega^{(t)}$ can be drawn with or without replacement. When dealing with MF models, instead of sub-sampling the data arbitrarily, one might come up with more clever sub-sampling schemas that could reduce the computational burden drastically by enabling parallelism. In the next section, we will describe our novel method, PSGLD, where we utilise a systematic sub-sampling schema by exploiting the conditional independence structure of MF models.

\section{Parallel SGLD for Matrix Factorisation}

In this section, we describe the details of PSGLD. Inspired by \cite{Liu2010,gemulla2011,recht2013parallel}, PSGLD utilises a biased sub-sampling schema where the observed data is carefully partitioned into mutually disjoint blocks and the latent factors are also partitioned accordingly. An illustration of this approach is depicted in Figure~\ref{fig:illus}. In this particular example, the observed matrix $\V$ is partitioned into $3 \times 3$ disjoint blocks and the latent factors $\W$ and $\Hm$ are partitioned accordingly into $3 \times 1$ and $1 \times 3$ blocks. At each iteration, PSGLD sub-samples $3$ blocks from $\V$, called as the \emph{parts}, in such a way that the blocks in a part will not `touch' each other in any dimension of $\V$, as illustrated in Figure~\ref{fig:illus}. This biased sub-sampling schema enables parallelism, since given a part, the SGLD updates can be applied to different blocks of the latent factors in parallel.

In the example given in Figure~\ref{fig:illus}, we arbitrarily partition the data into $9$ equal-sized blocks where these blocks are obtained in a straightforward manner by partitioning $\V$ using a $3\times 3$ grid. In the general case, the observed matrix $\V$ will be partitioned into $B \times B = B^2$ blocks and these blocks can be formed in a data-dependent manner, instead of using simple grids.

Let us formally define a \emph{block} and a \emph{part}. First, we need to define a partition of a set ${\cal S}$ as ${\cal P}_B({\cal S})$ where ${\cal P}_B({\cal S})$ contains non-empty disjoint subsets of ${\cal S}$, whose union is equal to ${\cal S}$. Here, $B$ denotes the number of subsets that the partition ${\cal P}$ contains. We will define the \emph{block}s and the \emph{part}s by using partitions of the sets $[I]$ and $[J]$. 

\begin{definition}
A block, $\Lambda \subset [I] \times [J]$ is the Cartesian product of two sets, one of them being in ${\cal P}_B([I])$ and the other one being in ${\cal P}_B([J])$. Formally, it is defined as follows:
\begin{align}
\Lambda =  {\cal I} \times {\cal J} 
\end{align}
where ${\cal I} \in {\cal P}_B([I])$ and ${\cal J} \in {\cal P}_B([J])$.
\end{definition}

\begin{definition}
A part, $\Pi^{(t)} \subset [I] \times [J]$ at iteration $t$, is a collection of mutually disjoint blocks and is defined as follows: 
\begin{align}
\Pi^{(t)} = \cup_{b = 1}^B \Lambda_b^{(t)} = \cup_{b = 1}^B {\cal I}_b^{(t)} \times {\cal J}_b^{(t)} 
\end{align}
where all the blocks $\Lambda_b^{(t)}$ are mutually disjoint, formally,
\begin{align*}
&{\cal I}_b^{(t)} \in {\cal P}_B([I]), \qquad
{\cal J}_b^{(t)} \in {\cal P}_B([J]) \\
&{\cal I}_b^{(t)} \cap {\cal I}_{b'}^{(t)} = \emptyset, \qquad
{\cal J}_b^{(t)} \cap {\cal J}_{b'}^{(t)} = \emptyset, \qquad \forall b \neq b' .
\end{align*}
\end{definition}
Suppose we read a part $\Pi^{(t)} = \cup_{b=1}^B \Lambda_b^{(t)}$ at iteration $t$. Then the SGLD updates for $\W$ can be written as follows:
\begin{align}
\nonumber \Delta \W^{(t)} =& \epsilon^{(t)} \Bigl(\frac{N}{|\Pi^{(t)}|} \hspace{-2pt} \sum_{(i,j)\in \Pi^{(t)}} \hspace{-10pt} \nabla_{\W}  \log p(v_{ij} |\cdot) \\
\nonumber     &+\nabla_{\W}  \log p(\W^{(t-1)})    \Bigr)  +\noisebw^{(t)}  \\
\nonumber =&  \epsilon^{(t)} \Bigl(  \frac{N}{|\Pi^{(t)}|} \hspace{-1pt} \sum_{b=1}^B \sum_{(i,j)\in \Lambda_b^{(t)}} \hspace{-10pt} \nabla_{\W}  \log p(v_{ij} | \cdot) \\
 &+ \nabla_{\W}  \log p(\W^{(t-1)})   \Bigr)  + \noisebw^{(t)} \label{eqn:psgld_1}
\end{align}
Since all $\Lambda_b^{(t)}$ are mutually disjoint, we can decompose Equation~\ref{eqn:psgld_1} into $B$ \emph{interchangeable} updates (i.e., they can be applied in any order), that are given as follows: $\W_b^{(t)} = \W_b^{(t-1)} + \Delta \W_b^{(t)}$, where
\begin{align}
\nonumber \Delta \W_b^{(t)} =& \epsilon^{(t)} \Bigl(  \frac{N}{|\Pi^{(t)}|} \hspace{-4.5pt} \sum_{(i,j)\in \Lambda_b^{(t)}} \hspace{-10pt}  \nabla_{\W_b}  \log p(v_{ij} | \W_b^{(t-1)},\Hm_b^{(t-1)})\\   
  & +\nabla_{\W_b}  \log p(\W_b^{(t-1)})\Bigr)  + \noisebw_b^{(t)}  \label{eqn:psgld_w} 
\end{align} 
for all $b = 1,\dots, B$. Here, $\W_b^{(t)}$ and $\Hm_b^{(t)}$ are the latent factor blocks at iteration $t$, that are determined by the current data block $\Lambda_b^{(t)} = {\cal I}_b^{(t)} \times {\cal J}_b^{(t)} $ and are formally defined as follows:
\begin{align*}
\W_b^{(t)} &\equiv \{ w_{ik}^{(t)} | i \in {\cal I}_b^{(t)}, k \in [K] \}\\
\Hm_b^{(t)} &\equiv \{ h_{kj}^{(t)} | j \in {\cal J}_b^{(t)}, k \in [K] \}
\end{align*}
The noise matrix $\noisebw_b^{(t)}$ is of the same size as $\W_b$ and its entries are independently Gaussian distributed with mean $0$ and variance $2 \epsilon^{(t)}$.

Similarly, we obtain $B$ interchangeable update rules for $\Hm$ that are given as follows: $\Hm_b^{(t)} = \Hm_b^{(t-1)} + \Delta \Hm_b^{(t)}$, where
\begin{align}
\nonumber \Delta \Hm_b^{(t)} =&  \epsilon^{(t)} \Bigl( \frac{N}{|\Pi^{(t)}|} \hspace{-2pt} \sum_{(i,j)\in \Lambda_b^{(t)}} \hspace{-10pt} \nabla_{\Hm_b}  \log p(v_{ij} | \W_b^{(t-1)},\Hm_b^{(t-1)}) \\
  & + \nabla_{\Hm_b}  \log p(\Hm_b^{(t-1)})   \Bigr)  +  \noisebh_b^{(t)} \label{eqn:psgld_h}
\end{align} 
for all $b = 1,\dots, B$. Similarly, $\noisebh_b^{(t)}$ is of the same size as $\Hm_b$ and its entries are independently Gaussian distributed with mean $0$ and variance $2 \epsilon^{(t)}$. The parallelism of PSGLD comes from the fact that all these $B$ update rules are interchangeable, so that we can apply them in parallel. The pseudo-code of PSGLD is given in Algorithm~\ref{algo:psgld}.

\subsection{Convergence Analysis}

Since we are making use of a biased sub-sampling schema, it is not clear that the samples generated by PSGLD will converge to the Bayesian posterior. In this section, we will define certain conditions on the selection of the parts and provided these conditions hold, we will show that the probability distribution of the samples $\W^{(t)}$ and $\Hm^{(t)}$ converges to the Bayesian posterior $p(\W,\Hm|\V)$. 

For theoretical use, we define $\thetab$ as the parameter vector, that contains both $\W$ and $\Hm$:
\begin{align}
\thetab \triangleq [ \textbf{vec}(\W)^\top, \textbf{vec}(\Hm)^\top ]^\top 
\end{align}
where $\textbf{vec}(\cdot)$ denotes the vectorisation operator. We also define
\begin{align}
\nonumber{\cal L}(\thetab^{(t)}) &\triangleq \log p(\thetab^{(t)}) + \hspace{-8pt} \sum_{i,j \in [I] \times [J]} \hspace{-5pt} \log p(v_{ij}|\thetab^{(t)})\\
\nonumber \hat{{\cal L}}(\thetab^{(t)}) &\triangleq  \log p(\thetab^{(t)}) + \frac{N}{|\Pi^{(t)}|} \sum_{i,j \in \Pi^{(t)}}  \log p(v_{ij}|\thetab^{(t)})
\end{align}
Then, the stochastic noise is given by
\begin{align}
\xib^{(t)} = \nabla_\thetab \hat{{\cal L}}(\thetab^{(t)}) - \nabla_\thetab {\cal L}(\thetab^{(t)}).
\end{align}
Under the following conditions Theorem~\ref{thm:conv} holds. 
\begin{condition}
The step size $\epsilon^{(t)}$ satisfies Equation~\ref{eqn:eps}. 
\end{condition}
\begin{condition}
The part $\Pi^{(t)}$ is chosen from $B$ nonoverlapping parts whose union covers the whole observed matrix $\V$ (e.g., the parts given in Figure~\ref{fig:illus}). The probability of choosing a part $\Pi^{(t)}$ at iteration $t$ is proportional to its size:
\begin{align*}
p(\Pi^{(t)} = \Pi) = \frac{|\Pi|}{N}.
\end{align*} 
\label{cond:parts}
\end{condition}
\begin{condition}
$\mathds{E}[(\xib^{(t)})^k] < \infty$, for integer $k \geq 2$.
\label{cond:sampling}
\end{condition}
\begin{thm}
\label{thm:conv}
Let $q_t(\thetab)$ be the probability density function of the samples $\thetab^{(t)}$ that are generated by PSGLD. Then, the probability distribution of $\thetab^{(t)}$ converges to the Bayesian posterior $p(\thetab|\V)$:
\begin{align}
\lim_{t \rightarrow \infty} q_t(\thetab) = p(\thetab|\V).
\end{align}
\end{thm}
\begin{proofsketch}[Proof sketch]
Under Condition~\ref{cond:parts}, we can show that $\hat{{\cal L}}$ is an unbiased estimator of ${\cal L}$; therefore the stochastic noise $\xib^{(t)}$ is zero-mean:
\begin{align*}
\mathds{E}[\xib^{(t)}] = 0.
\end{align*}
The rest of the proof is similar to \cite{icml2014c2_satoa14}. Under conditions 1 and 3, we can show that $q_t(\thetab)$ follows the (multi-dimensional) Fokker-Plank equation and therefore the stationary distribution of $q_t(\thetab)$ is $p(\thetab|\V) \propto \exp(-{\cal L}(\thetab)) $. 
\end{proofsketch}

\begin{algorithm2e}[t]
\KwIn{\hspace{7pt} $\V$, $\W^{(0)}$, $\Hm^{(0)}$, $T$, $B$, ${\cal P}_B([I])$, ${\cal P}_B([J])$}
\KwOut{ $ \{ \W^{(t)},\Hm^{(t)} \}_{t=1}^T$ }

\For{ $t \leftarrow 1$ to $T$ }{
Set the step size $\epsilon^{(t)}$\\
Pick a part $\Pi^{(t)} = \cup_{b=1}^B \Lambda_b^{(t)}$\\
\ForPar{ each block $\Lambda_b^{(t)}$ in $\Pi^{(t)}$ }{
	$\W_b^{(t)}  = \W_b^{(t-1)} + \Delta \W_b^{(t)}$ \hfill \textit{ \small \color{magenta} //(Eq. \ref{eqn:psgld_w})}\\
	$\Hm_b^{(t)} \hspace{3pt} = \Hm_b^{(t-1)} + \Delta \Hm_b^{(t)}$  \hfill \textit{ \small \color{magenta} //(Eq. \ref{eqn:psgld_h})}\\
	\textit{\small \color{magenta} /* Optional mirroring step for non-negativity.}\\
	\textit{\small \color{magenta} $|\cdot|$ is element-wise absolute value operator. */} \\
	$\W_b^{(t)} \leftarrow  |\W_b^{(t)}| $\\
	$\Hm_b^{(t)} \hspace{3pt} \leftarrow  |\Hm_b^{(t)}| $
}
}
  \caption{PSGLD for matrix factorisation.}
  \label{algo:psgld}
\end{algorithm2e}

\subsection{Non-negativity Constraints}

In certain applications, all the elements of $\V$, $\W$, and $\Hm$ are required to be non-negative, that is known as the \emph{non-negative matrix factorisation} (NMF) \cite{lee99}. As we will illustrate in Section~\ref{sec:experiments}, the non-negativity constraint is often a necessity in certain probabilistic models, where we essentially decompose the parameters of the probabilistic model that are non-negative by definition (e.g., the intensity of a Poisson distribution or the mean of a gamma distribution).

In an SGD framework, the latent factors can be kept in a constraint set by using projections that apply the minimum force to keep the variables in the constraint set. However, since we are in an MCMC framework, it is not clear that appending a projection step to the PSGLD updates would still result in a proper MCMC method. Instead, similar to \cite{PatTeh2013a}, we make use of a simple {mirroring trick}, where we replace the negative entries of $\W^{(t)}$ and $\Hm^{(t)}$ with their absolute values. Formally, we let $w_{ik}$ and $h_{kj}$ take values in the whole $\mathds{R}$, however we parametrise the prior and the observation models with the absolute values, $|w_{ik}|$ and $|h_{kj}|$. Since $w_{ik}^{(t)}$ and $-w_{ik}^{(t)}$ (similarly, $h_{kj}^{(t)}$ and $-h_{kj}^{(t)}$) will be equiprobable in this setting, we can replace the negative elements of $\W^{(t)}$ and $\Hm^{(t)}$ with their absolute values without violating the convergence guarantee.

\section{Experiments}
\label{sec:experiments}

In this section we will present our experiments where we evaluate PSGLD on both synthetic and real datasets using the \emph{non-negative matrix factorisation} (NMF) model. In order to be able to cover a wide range of likelihood functions, we consider the following probabilistic model:
\begin{gather}
\nonumber p(\W) = \prod_{ik} {\cal E}(w_{ik}; \lambda_w) , \qquad
\nonumber p(\Hm) = \prod_{kj} {\cal E}(h_{kj}; \lambda_h) \\
p(\V|\W\Hm) = \prod_{ij} {\cal TW}(v_{ij}; \sum_{k} w_{ik} h_{kj}, \phi, \beta ) \label{eqn:tweedienmf}
\end{gather}
where $\V \in \mathds{R}_+^{I \times J}$, $\W \in \mathds{R}_+^{I \times K}$, and $\Hm \in \mathds{R}_+^{K \times J}$. Here, ${\cal E}$ and ${\cal TW}$ denote the exponential and Tweedie distributions, respectively. The Tweedie distribution is an important special case of the exponential dispersion models \cite{jorgensen1997} and has shown to be useful for factorisation models \cite{yilmazGTF}. The Tweedie density can be written in the following form:
\begin{align*}
	{\cal TW}(v; \mu, \phi,\beta) = \frac1{K(x,\phi,\beta)} \exp \Bigl( - \frac1{\phi} d_\beta (v || \mu )  \Bigr) 
\end{align*}
where $\mu$ is the mean, $\phi$ is the dispersion (related to the variance), $\beta$ is the power parameter, $K(\cdot)$ is the normalizing constant, and $d_\beta(\cdot)$ denotes the $\beta$-divergence that is defined as follows:
\begin{align*}
 d_\beta(v|| \mu )  =  \frac{v^{\beta}}{\beta(\beta-1)} - \frac{v \mu^{\beta-1}}{\beta-1} + \frac{\mu^{\beta}}{\beta}.
\end{align*}
The $\beta$-divergence generalises many divergence functions that are commonly used in practice. As special cases, we obtain the Itakura-Saito divergence, Kullback-Leibler divergence, and the Euclidean distance square, for $\beta=0,1,2$, respectively. From the probabilistic perspective, different choices of $\beta$ yield important distributions such as gamma ($\beta=0$), Poisson ($\beta=1$), Gaussian ($\beta=2$), compound Poisson ($0<\beta<1$), and inverse Gaussian ($\beta=-1$) distributions. Due to a technical condition, no Tweedie model exists for the interval $1<\beta<2$, but for all other values of $\beta$, one obtains the very rich family of Tweedie stable distributions \cite{jorgensen1997}. Thanks to the flexibility of the Tweedie distribution, we are able to choose an observation model by changing a single parameter $\beta$, without modifying the inference algorithm.

In most of the special cases of the Tweedie distribution, the normalizing constant $K(\cdot)$ is an infinite sum and cannot be written in a simple analytical form. Fortunately, provided that $\phi$ and $\beta$ are given, the normalizing constant becomes irrelevant since it does not depend on the mean parameter $\mu$ and therefore $\W$ and $\Hm$. Consequently, the PSGLD updates only involve the partial derivatives of the $\beta$-divergence with respect to $w_{ik}$ and $h_{kj}$, which is tractable.

\begin{figure*}[t]
\begin{center}
 \subfigure[]{\includegraphics[width=0.5\columnwidth]{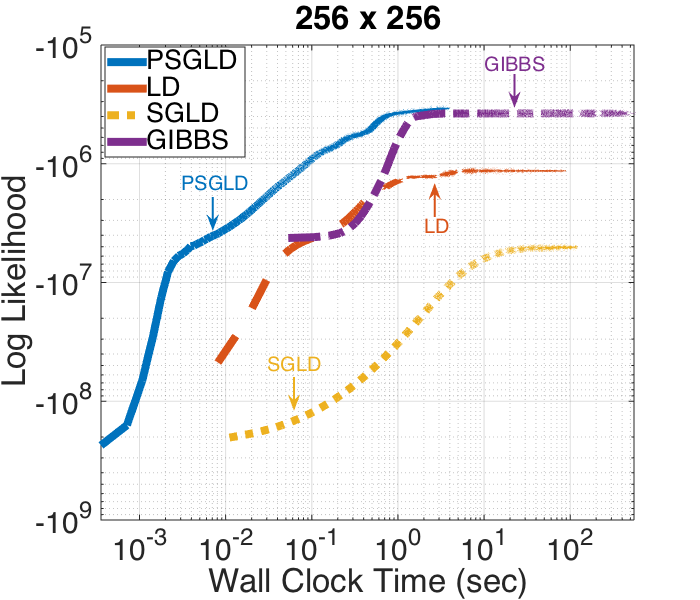}
\includegraphics[width=0.5\columnwidth]{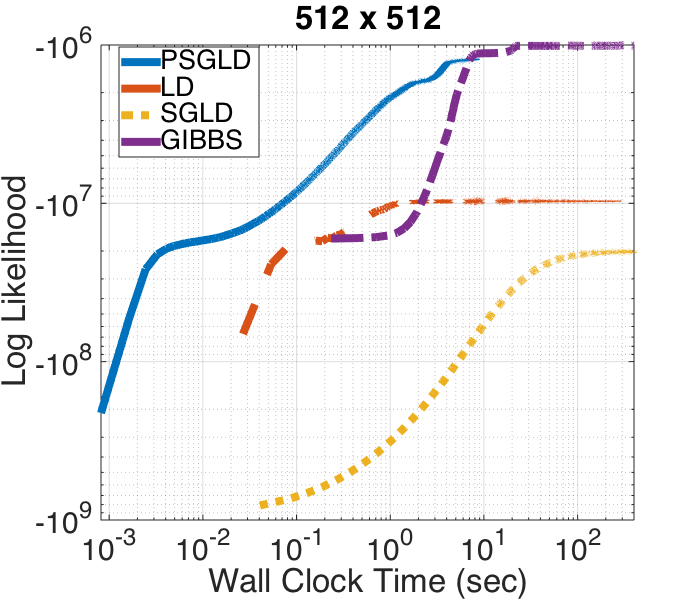}
\includegraphics[width=0.5\columnwidth]{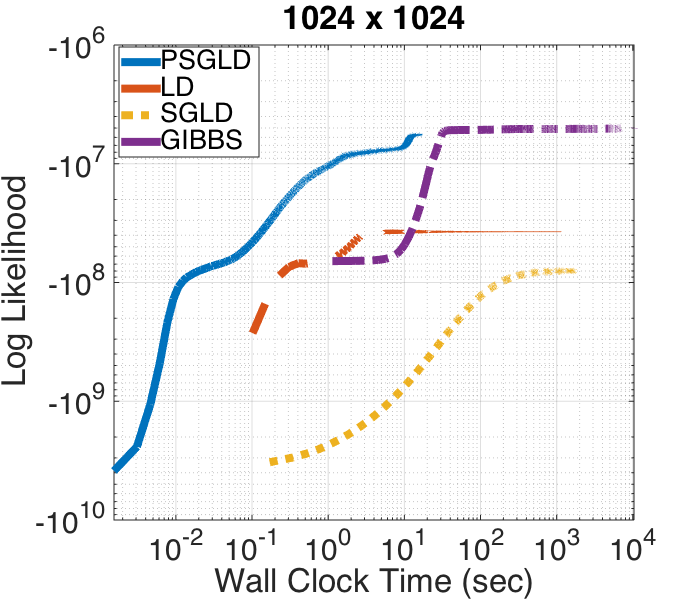} \label{fig:sm_pois}}
 \subfigure[]{\includegraphics[width=0.5\columnwidth]{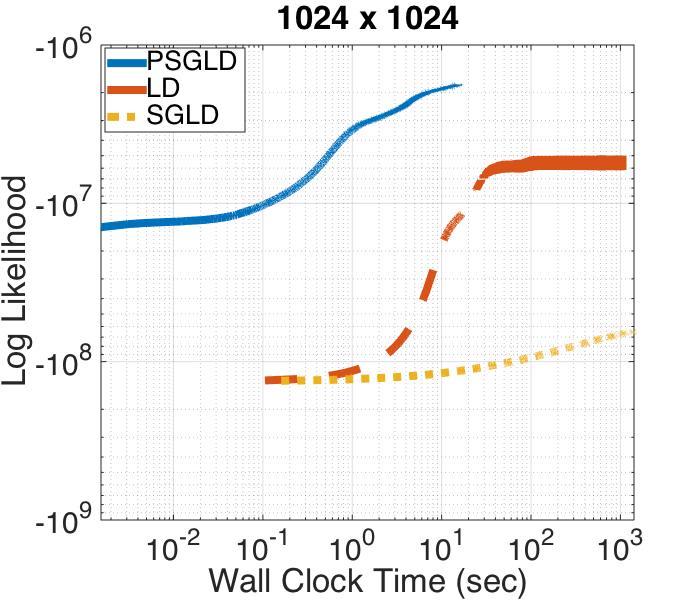} \label{fig:sm_cp}}
\vspace{-10pt}
\caption{Shared-memory experiments with a) the Poisson observation model b) the compound Poisson observation model.}
\label{fig:sm}
\end{center}
\vskip -0.2in
\end{figure*} 

\subsection{Experimental Setup}

We will compare PSGLD with different MCMC methods, namely the Gibbs sampler, LD, and SGLD. We will conduct our experiments in two different settings: 1) a shared-memory setting where the computation is done on a single multicore computer 2) a distributed setting where we make use of a cluster of computing nodes\footnote{For the source code for both settings (CUDA and OpenMPI), please contact the authors.}. 

It is easy to derive the update equations required by the gradient-based methods for the Tweedie-NMF model. However, developing a Gibbs sampler for this general model is unfortunately not obvious. We could derive Gibbs samplers for certain special cases of the Tweedie model, such as the Poisson-NMF \cite{cemgil:cin} where $\beta = 1$ and $\phi = 1$. Moreover, in order the full conditional distributions that are required by the Gibbs sampler, we need to introduce an auxiliary tensor and augment the probabilistic model in Equation~\ref{eqn:tweedienmf} as follows:
\begin{gather*}
\nonumber p(w_{ik}) = {\cal E}(w_{ik}; \lambda_w) , \qquad
\nonumber p(h_{kj}) =  {\cal E}(h_{kj}; \lambda_h) \\
 p(s_{ijk}) = {\cal PO}(s_{ijk}; w_{ik} h_{kj}), \quad
v_{ij} =  \sum_{k} s_{ijk} 
\end{gather*} 
where ${\cal PO}$ denotes the Poisson distribution. 

The LD and Gibbs samplers require to pass on the whole observed matrix $\V$ at every iteration. The Gibbs sampler further requires the whole auxiliary tensor $\mathbf{S} \equiv \{s_{ijk}\} \in \mathds{R}^{I\times J \times K}$ to be sampled at each iteration. 

\subsection{Shared-Memory Setting}

In this section, we will compare the mixing rates and the computation times of all the aforementioned methods in a shared-memory setting. We will first compare the methods on synthetic data, then on musical audio data.

We conduct all the shared-memory experiments on a MacBook Pro with $2.5$GHz Quad-core Intel Core i7 CPU, $16$ GB of memory, and NVIDIA GeForce GT 750M graphics card. We have implemented PSLGD on the GPU in CUDA C. We have implemented the other methods on the CPU in C, where we have used the GNU Scientific Library and BLAS for the matrix operations. 

\subsubsection{Experiments on Synthetic Data}

In order to be able to compare all the methods, in our first experiment we use the Poisson-NMF model. We first generate $\W$, $\Hm$, and $\V$ by using the generative model. Then, we run all the methods in order to obtain the samples $\{\W^{(t)}, \Hm^{(t)}\}_{t=1}^T$. For simplicity, we choose $I = J$ and we set $K = 32$. In order to obtain the blocks, we partition the sets $[I]$ and $[J]$ into $B = I/32$ equal pieces, where we simply partition $\V$ by using a $B \times B$ grid, similar to the example given in Figure~\ref{fig:illus}. Initially, we choose $B$ different parts whose union cover the whole observed matrix $\V$, similar to the ones in Figure~\ref{fig:illus}. At each iteration, we choose one of these parts in cyclic order, i.e. we proceed to the next part at each iteration and return the first part after iteration $Bk$ with integer $k \geq 1$. Since the sizes of all the parts are the same, Condition~\ref{cond:parts} is satisfied.

In LD, we use a constant step size $\epsilon$, whereas in SGLD and PSGLD, we set the step sizes as $\epsilon^{(t)} = (a/t)^b$, where $b \in (0.5,1]$. For each method, we tried several values for the parameters and report the results for the best performing ones. In LD we set $\epsilon = 0.2$, in SGLD we set $a=1$, $b = 0.51$, and in PSGLD we set $a=0.01$ and $b = 0.51$. The results are not very sensitive to the actual value of $a$ and $b$, provided these are set in a reasonable range. Furthermore, in SGLD, we draw the sub-samples $\Omega^{(t)}$ with a with-replacement manner, where we set $|\Omega^{(t)}| = IJ/32$. 

Figure~\ref{fig:sm_pois} shows the mixing rates and the running times of the methods under the Poisson model for different data sizes. While plotting the log-likelihood of the state of the Markov chain is not necessarily an indication of convergence to the stationary distribution, nevertheless provides a simple indicator if the sampler is stuck around a low probability mode. 
We set the number of rows $I = 256$, $512$, $1024$ and we generate $T=10000$ samples from the Markov chain with each method. We can observe that, in all cases, SGLD achieves poor mixing rates due to the with-replacement sub-sampling schema while LD achieves better mixing rates than SGLD. Moreover, while the LD updates can be implemented highly efficiently using BLAS, the reduced data access of SGLD does not reflect in reduced computation time due to the random data access pattern when selecting sub-samples from $\V$. 

The results show that PSGLD and the Gibbs sampler seem to achieve much better mixing rates. However, we observe an enormous difference in the running times of these methods -- PSGLD is $700+$ times faster than the Gibbs sampler on a GPU, while achieving virtually the same quality. For example, in a model with $I = 1024$ rows, the Gibbs sampler runs for more than $3$ hours while PSGLD completes the burn-in phase in nearly $1$ second and generates $10$K samples from the Markov chain in less than $15$ seconds, even when there are more than $1$ million entries in $\V$. Naturally, this gap between PSGLD and the Gibbs sampler becomes more pronounced with increasing problem size. We also observe that PSGLD is faster than LD and SGLD by $60+$ folds while achieving a much better mixing rate.

\begin{figure*}[t]
\begin{center}
 \subfigure[]{\includegraphics[width=0.96\columnwidth]{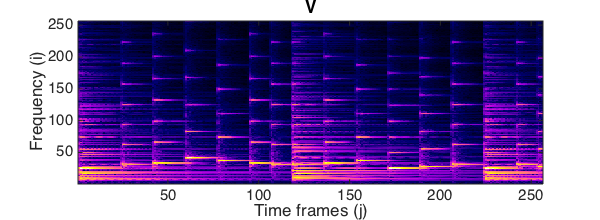} \label{fig:audiov}}
 \subfigure[]{\includegraphics[width=0.48\columnwidth]{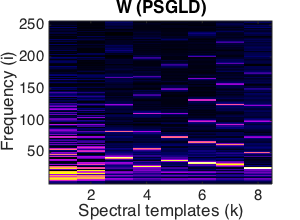}
\includegraphics[width=0.48\columnwidth]{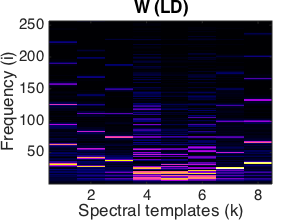}
\label{fig:audiow}}
\vspace{-15pt}
\caption{a) The audio spectrum of a short piano piece b) The spectral dictionaries learned by PSGLD and LD.}
\label{fig:audio}
\end{center}
\vskip -0.2in
\end{figure*}

We also evaluate PSGLD with ${\cal TW}(v;\mu,\phi=1,\beta =0.5)$ observation model, which corresponds to a compound Poisson distribution. This distribution is particularly suited for sparse data as it has a non-zero probability mass on $v=0$ and a continuous density on $v > 0$ \cite{jorgensen1997}. Even though the probability density function of this distribution cannot be written in closed-form analytical expression, fortunately we can still generate random samples from the distribution in order to obtain synthetic $\V$.

Since deriving a Gibbs sampler for the compound Poisson model is not obvious, we will compare only LD, SGLD, and PSGLD on this model. Figure~\ref{fig:sm_cp} shows the performance of these methods for $I = J = 1024$. We obtain qualitatively similar results; PSGLD achieves a much better mixing rate and is much faster than the other methods.

\subsubsection{Experiments on Audio}

The Tweedie-NMF model has been widely used for audio and music modelling \cite{fevotte2011algorithms}. In musical audio context, the observed matrix $\V$ is taken as a \emph{time-frequency} representation of the audio signal, such as the power or magnitude spectra that are computed via short-time Fourier transform. Here, the index $i$ denotes the \emph{frequency bins}, whereas the index $j$ denotes the \emph{time-frames}. An example audio spectrum belonging to a short piano excerpt ($5$ seconds) is given in Figure~\ref{fig:audiov}.    

When the audio spectrum $\V$ is decomposed by using an NMF model, each column of $\W$ will contain a different \emph{spectral template} and each row of $\Hm$ will contain the \emph{activations} through time for a particular spectral template. In music processing applications, each spectral template is expected to capture the spectral shape of a certain musical note and the activations are expected to capture the loudness of the notes. 

We decompose the audio spectrum given in Figure~\ref{fig:audiov} and visually compare the dictionary matrices that are learned by LD and PSGLD. The size of $\V$ is $I = J = 256$ and we set $K = 8$. For PSGLD, we partition the sets $[I]$ and $[J]$ into $B = 8$ equal pieces and we choose the parts in cyclic order at each iteration. With each method, we generate $10000$ samples but discard the samples in the burn-in phase ($5000$ samples). Figure~\ref{fig:audiow} shows the Monte Carlo averages that are obtained by different methods. We observe that PSGLD successfully captures the spectral shapes of the different notes and the chords that occur in the piece, even though the method is completely unsupervised. We also observe that LD is able to capture the spectral shapes of most of the notes as well, and estimates a less sparse dictionary. Furthermore, PSGLD runs in a much smaller amount of time; the running times of the methods are $3.5$ and $81$ seconds respectively for PSGLD and LD -- as a reference the Gibbs sampler needs to run for $533$ seconds on the same problem.

\begin{figure}[t]
\begin{center}
\includegraphics[width=0.99\columnwidth]{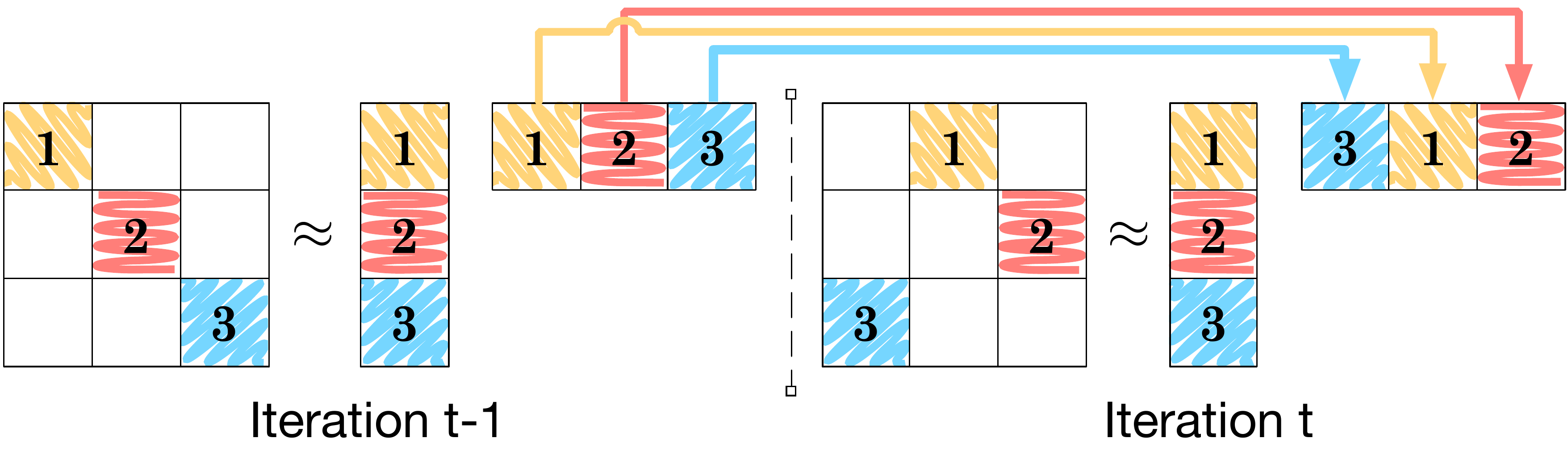}
\vspace{-10pt}
\caption{Illustration of the communication mechanism. There are $3$ nodes and $B = 3$ is selected as the number of nodes. The numbers inside the blocks denote the nodes in which the corresponding blocks are located. At each iteration, node $n$ transfers its $\Hm_b$ block to node $(n \mod B) +1$. The blocks $\W_b$ are kept in the same node throughout the process. This strategy implicitly determines the part to be used at the next iteration. }
\label{fig:comm}
\end{center}
\vskip -0.2in
\end{figure} 

\subsection{Distributed-Hybrid Setting}
 
In this section, we will focus on the implementation of PSGLD in a distributed setting, where each block of $\V$ might reside at a different node. We will consider a distributed architecture that contains three main components: 1) the data nodes that store the blocks of $\V$ 2) the computational nodes that execute the PSGLD updates 3) the main node that is only responsible for submitting the jobs to the computational nodes only at the beginning of the sampling process.

In the distributed setting, we implement PSGLD by a message passing protocol in C using the OpenMPI library. PSGLD is naturally suited for message passing environments, and the low level control on the distributed computations provide more insight than other platforms such as Hadoop MapReduce. On the other hand, it is straightforward to implement PSGLD in a MapReduce environment for commercial and fault-tolerant applications.

In our implementation, we make use of an efficient communication mechanism, where we set the number of blocks $B$ to the number of available nodes. As illustrated in Figure~\ref{fig:comm}, throughout the sampling process, each node is responsible only for a certain $\W_b$ block; however, at the end of each iteration it transfers the corresponding $\Hm_b$ block to its adjacent node in a cyclic fashion. With this mechanism, the part $\Pi^{(t)}$ is determined implicitly at each iteration depending on the current locations of the factor blocks $\W_b$ and $\Hm_b$. Besides, as opposed to many distributed MCMC methods such as DSGLD, this mechanism enables PSGLD to have a much lower communication cost, especially for large $I$, $J$, and $B$ values.

We conduct our distributed-setting experiments on a cluster with $15$ computational nodes where each computational node has $8$ Intel Xeon $2.50$GHz CPUs and $16$ GB of memory. Therefore, provided that the memory is sufficient, we are able to run $120$ concurrent processes on our cluster. In our experiments, by assuming that the network connection between the computational nodes is sufficiently fast, we will assume that we have at most $120$ computational nodes.

\begin{figure}
\begin{center}
\includegraphics[width=\columnwidth]{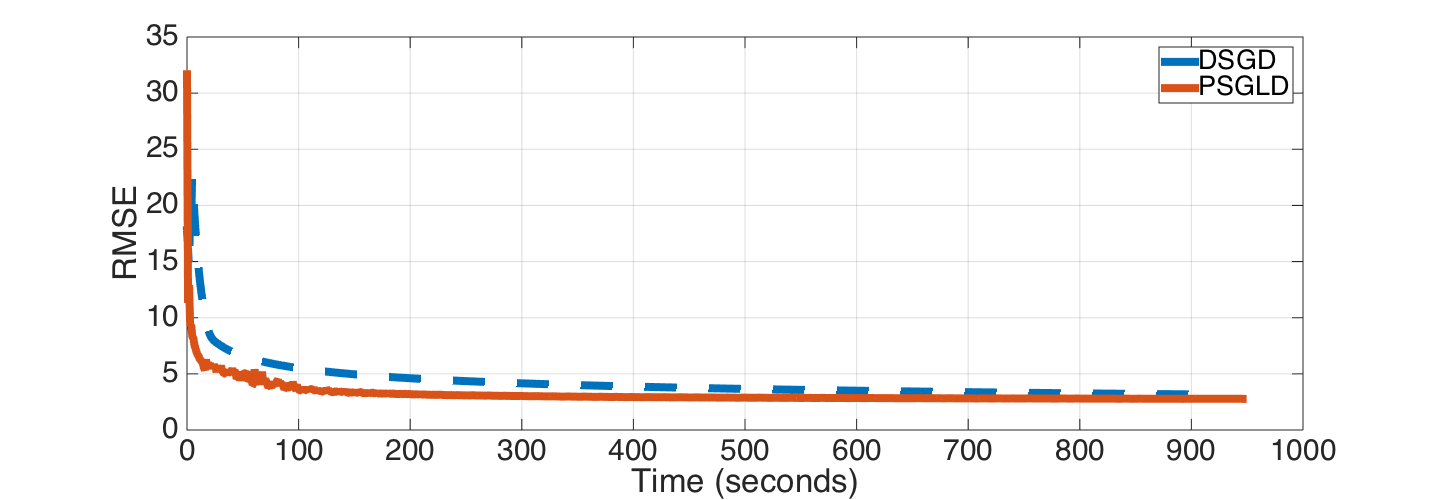}
\vspace{-13pt}
\caption{RMSE values on MovieLens 10M dataset.}
\label{fig:rmse}
\end{center}
\vskip -0.2in
\end{figure}

We evaluate PSGLD on a large movie ratings dataset, MovieLens $10$M (\url{grouplens.org}). This dataset contains $10$ million ratings applied to $I = 10681$ movies by $J = 71567$ users, resulting in a sparse $\V$ where $1.3\%$ of $\V$ is non-zero. In all our experiments, we set $K = 50$, $\beta = \phi = 1$, and we set $B$ to the number of available nodes where we partition the sets $[I]$ and $[J]$ into $B$ equal pieces similar to the shared-memory experiments. In these experiments, the sizes of the parts are close to each other, therefore our communication mechanism satisfies Condition $2$.

In our first experiment, our goal is to contrast the speed of our sampling algorithm to a distributed optimisation algorithm. Clearly, the goals of both computations are different (a sampler does not solve an optimisation problem unless techniques such as simulated annealing is being used), yet monitoring the root mean squared error (RMSE) between $\V$ and $\W\Hm$ throughout the iterations provides a qualitative picture about the convergence behaviour of the algorithms. 
Figure~\ref{fig:rmse} shows the RMSE values of PSGLD and the distributed stochastic gradient descent (DSGD) algorithm \cite{gemulla2011} for $1000$ iterations with $B = 15$. We observe that a very similar convergence behaviour and the running times for both methods. The results indicate that, PSGLD makes Bayesian inference possible for MF models even for large datasets by generating samples from the Bayesian posterior, while at the same time being as fast as the state-of-the-art distributed optimisation algorithms.

\begin{figure}[t]
\begin{center}
\subfigure[]{\includegraphics[width=0.545\columnwidth]{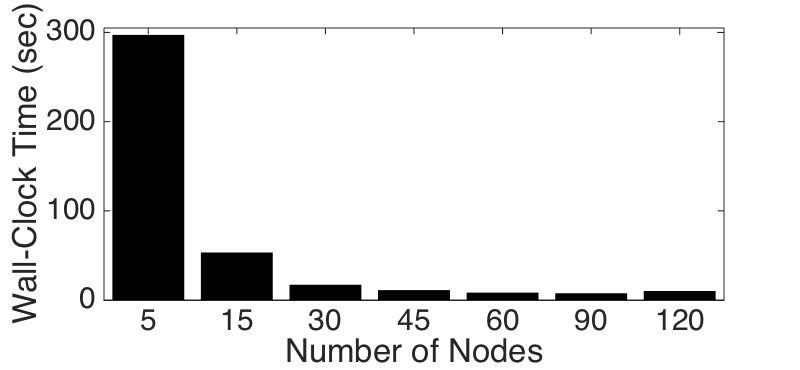} \label{fig:timings}}
\subfigure[]{\includegraphics[width=0.416\columnwidth]{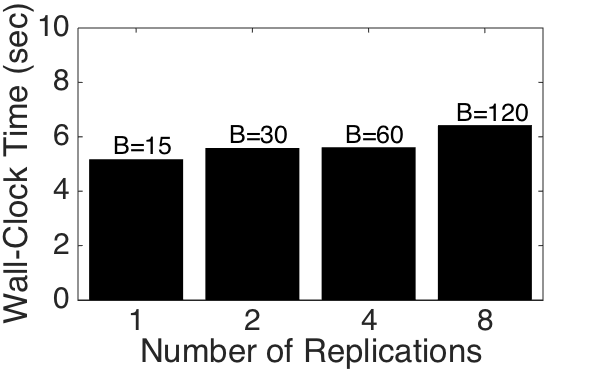} \label{fig:scale}}
\vspace{-17pt}
\caption{Scalability of PSGLD. a) The size of the data is kept fixed, the number of nodes is increased b) The size of the data and the number of nodes are increased proportionally.}
\end{center}
\vskip -0.2in
\end{figure}

In our last set of experiments, we demonstrate the scalability of PSGLD. Firstly, we differ the number of nodes from $5$ to $120$ and generate $100$ samples in each setting. Figure~\ref{fig:timings} shows the running times of PSGLD for different number of nodes. The results show that, the running time reduces almost quadratically as we increase the number of nodes until $B=90$. For $B=120$, the communication cost dominates and the running time increases.

Finally, in order to illustrate how PSGLD scales with the size of the data, we increase the size of $\V$ while increasing the number of nodes accordingly. We start with the original dataset and $15$ nodes, then we duplicate $\V$ in both dimensions (the number of elements quadruples) and set the number of nodes to $30$. We repeat this procedure two more times, where the ultimate dataset becomes of size $683.584 \times 4.580.288$ with $640$ million non-zero entries and the number of nodes becomes $120$. Figure~\ref{fig:scale} shows the running times of PSGLD with $T = 10$ for increasing data sizes and number of nodes. The results show that, even though we increase the size of the data $64$ folds, the running time of PSGLD remains nearly constant provided we can increase the number of nodes proportionally.

\section{Conclusion}
We have described a scalable MCMC method for sampling from the posterior distribution of a MF model and tested the performance of our approach in terms of accuracy, speed and scalability on various modern architectures. Our results suggest that, contrary to the established folklore in ML, inference methods for `big data' are not limited to optimisation, and Monte Carlo methods are as competitive in this regime as well. The existence of efficient samplers paves the way to full Bayesian inference; due to lack of space we have not presented natural applications such as model selection.

We conclude with the remark that it is rather straightforward to extend PSGLD to more structured models such as 
coupled matrix and tensor factorisation models. Here, several datasets are decomposed simultaneously and the distributed nature of PSGLD is arguably even more attractive when data are naturally distributed to different physical locations. 

\section*{Acknowledgments}
This work is supported by the Scientific and Technological Research Council of Turkey (T\"{U}B\.{I}TAK) Grant no. 113M492. Umut \c Sim\c sekli is also funded by a PhD Scholarship from T\"{U}B\.{I}TAK.

\bibliography{psgld_nmf}
\bibliographystyle{icml2015}

\end{document}